\documentclass[twoside]{article}
\usepackage[T1]{fontenc}
\usepackage[accepted]{aistats2021}
\usepackage{tikz}
\usetikzlibrary{bayesnet}
\usepackage{shortbold}
\usepackage{url}
\usepackage{subcaption}
\usepackage[colorinlistoftodos,textsize=tiny, textwidth=18mm]{todonotes}
\usepackage{amsmath,amssymb}
\usepackage[ruled,vlined]{algorithm2e}
\usepackage{color, colortbl}
\definecolor{LightCyan}{rgb}{0.88,1,1}
\usepackage{multirow}

\newif\ifincludecomment
\includecommenttrue 
\ifincludecomment
  
\else
  \NewEnviron{guidance}{}{}
\fi
\ifincludecomment
\newcommand{\maybecomment}[1]{\todo[color=olive!40]{#1}} 
\newcommand{\maybetohere}[1]{\todo[color=red!40]{#1}} 
\newcommand{\maybedelete}[1]{\todo[color=blue!40]{#1}} 

\else
  \newcommand{\maybecomment}[1]{}
  
\newcommand{\maybedelete}[1]{} 
\fi
\newcommand{\amostohere}[1]{{\color{black}\maybetohere{AMOS HERE}}}
\usepackage[accepted]{aistats2021}
\usepackage[round]{natbib}

\begin{document}



\twocolumn[

% \aistatstitle{Neural Networks exhibit confounder and collider bias: mitigation via latent adversarial examples}

\aistatstitle{Latent Adversarial Debiasing: Mitigating Collider Bias in Deep Neural Networks}

\aistatsauthor{ Luke Darlow \And Stanisław Jastrzębski \And  Amos Storkey }

% Like in https://arxiv.org/pdf/1106.4509.pdf
\aistatsaddress{ School of Informatics, \\ University of Edinburgh \And Jagiellonian University \& \\ Molecule.one \And School of Informatics, \\
  University of Edinburgh }

  ]

% \aistatsaddress{ School of Informatics \\
%   University of Edinburgh \\
%   10 Crichton St, Edinburgh EH8 9AB \\ \And Jagiellonian University \\ Cracow,Poland \& \\ Molecule.one \And School of Informatics \\
%   University of Edinburgh \\
%   10 Crichton St, Edinburgh EH8 9AB \\ } 

%   ]

\begin{abstract}
  Collider bias is a harmful form of sample selection bias that neural networks are ill-equipped to handle. This bias manifests itself when the underlying causal signal is strongly correlated with other confounding signals due to the training data collection procedure. In the situation where the confounding signal is easy-to-learn, deep neural networks will latch onto this and the resulting model will generalise poorly to in-the-wild test scenarios. We argue herein that the cause of failure is a combination of the deep structure of neural networks and the greedy gradient-driven learning process used -- one that prefers easy-to-compute signals when available. We show it is possible to mitigate against this by generating bias-decoupled training data using latent adversarial debiasing (LAD), even when the confounding signal is present in 100\% of the training data. By training neural networks on these adversarial examples, we can improve their generalisation in collider bias settings. Experiments show state-of-the-art performance of LAD in label-free debiasing with gains of 76.12\% on background coloured MNIST, 35.47\% on foreground coloured MNIST, and 8.27\% on corrupted CIFAR-10. 
\end{abstract}

\section{Introduction}
Invariably, in real-world machine learning settings, training and test sets are different. This general phenomenon has become known as \emph{dataset shift}~\citep{caron_deep_2018}. Yet there are many causes for such shifts~\citep{storkey_when_2009}. One common scenario is \emph{sample selection bias} \citep{heckman_sample_1979} where the process of curating a training dataset differs from the process by which data arrives during deployment. This issue is ubiquitous; even standard machine learning benchmarks (e.g. ImageNet) contain images selected for the clarity with which a class is represented, a clarity missing in many real applications, where for example you might see an object occluded.

One pernicious form of sample selection bias is \emph{collider bias}. This is illustrated and characterised in Figure~\ref{fig:GM}. Consider two variables: a \emph{causal variable} that determines the target, and what we will call a \emph{confounding variable}, that is not directly related to the target. In collider bias, these two variables that are, for the most part, independent in test scenario, become co-dependent in the training sample because of the restrictive way the training data is selected. Collider bias can cause a predictive algorithm to mistakenly target information from features associated with the confounding variable rather than the causal variable; such features then do not generalise to the test scenario. 

The data processing inequality implies that information about the target from the causal signal must be greater than that of the confounding signal. However the confounding signal can be easier to discover than the causal signal, due to its ease of compute (a highly linear confounder, for example can be discovered before a highly non-linear causal relationship). When this is combined with the greediness of neural network learning, it can mean the causal signal is just never learnt about. It is precisely this scenario that is the topic of this paper.

We argue that collider bias can be a pervasive cause of non-robustness in deep neural networks (DNNs). Our main contribution is the demonstration of a specific approach to mitigate the situation: Latent Adversarial Debiasing (\textbf{LAD}), which pushes a network to recognise all sources of information for a problem by augmenting training using adversarially perturbed latent representations. In these latent representations, easy-to-learn confounding signals are decoupled from the classification targets, forcing networks to also learn information from the causal signal.

% In this paper, we illustrate that collider and confounder biases are a significant cause of non-robustness in deep neural networks (DNNs). We show that DNNs magnify the impact of these biases due to a combination of three things: their deep structure, the gradientg-driven learning process used, and the fact that non-causal variables are often more simply expressed in the data, making it easier for learning to latch onto the wrong signal. Our main contribution is an approach to mitigate the situation: \textbf{L}atent \textbf{A}dversarial \textbf{D}ebiasing (\textbf{LAD}), a method to counter over-simplification by augmenting the training data, by adversarially perturbing latent representations, such that easy-to-learn confounding signals are decoupled from the classification targets.

% In this paper, we illustrate that collider bias is a significant issue in real world settings, and indeed it is common cause of what is identified as non-robustness in deep neural networks.
% We show that neural networks are commonly fooled by situations of collider bias, due to the combination of their deep structure and the gradient-driven learning process used. We characterise the fundamental limitations in identifying the best predictor when collider bias is present, but suggest two approachers to mitigating the situation: countering over-simplification and using unsupervised information. We demonstrate the performance of these methods on related benchmarks, and realistic settings.

\subsection{Collider bias}
Consider a toy classification problem: distinguishing dog images from cat images. A collected training dataset for dogs and cats may be biased: people take pictures of dogs outside while they take them for a walk e.g. in a field, and take pictures of cats in their homes, e.g. on a sofa. Yet both cats and dogs go inside and outside.

It might be critical in the test setting to distinguish between dogs and cats each presented in indoor and outdoor settings.  In this training dataset the simple feature of the background colour in the image is a strong confounder for the real signal that needs to be detected -- the difference in appearance between dogs and cats.

We call the easy-to-learn bias-inducing signal the \textbf{confounding signal}, as opposed to the true \textbf{causal signal}. A model relying on the causal signal will generalise well, while a model relying on the confounding signal will generalise poorly. We propose a solution that hinges on the assumption that confounding signals are typically \emph{easier-to-compute} in that their gradients during learning are stronger than the gradients of causal signals (see Section \ref{sec:versus}). A similar assumption was also made in related earlier work \citep{nam_learning_2020,bahng_learning_2020,bras_adversarial_2020,minderer_automatic_2020}. We hypothesise that a specific form of adversarial examples can be used to augment training data such that confounding signal is decoupled from the causal signal. We show that adversarial data can be generated by gradient descent in the latent space of an autoencoder-like model to produce augmented training data. The effect is a reduced association between confounding and causal signals, requiring a model learning on this augmented data to rely on the causal signal. Our method, LAD, shows marked improvement over state-of-the-art, without relying on any presence of bias-free data (opposed to \citet{nam_learning_2020,bras_adversarial_2020} and \citet{bahng_learning_2020}).

% Our approach has a number of advantages:
% Our main contributions are as follows:

% The alterations are limited by an adversarial walk in a latent space that is prone to change easier-to-compute features (biases) first. 

% \begin{enumerate}
%     \item Marked improvement over state-of-the-art dataset debiasing, without relying on any presence of bias-free data (opposed to \cite{nam2020learning,bras2020adversarial});
%     \item A natural visualisation mechanism (via adversarial examples), allowing for qualitative diagnosis of what the model is relying on for decision making;
%     % \item A solution formulation that accounts for computational `ease', for which we argue is a fundamental component for dealing with biases that dominate learning.
% \end{enumerate}

\section{Problem Definition}\label{sec:problem}
% \textbf{}

% Commented out as it was repeating intordcution
% Consider the following toy scenario: you are an app developer and you want to build a camera-based app that can tell the difference between images of cats and images of dogs. You do not have the time to take the photos of cats and dogs you need to train a model, so you hire company A to do this for you. Unfortunately, company A gives this task to a single employee who happens to be allergic to cats. All the photos of cats are from a distance and are often occluded, while all the photos of dogs are close-up and clear. You train your model on this data but soon realise (when testing) that your model thinks all far-away dogs are cats and all close-up cats are dogs. This is because \emph{any image your model was trained on was causally influenced by two variables}: animal type and distance. Even though those two variables are not necessarily related, the sample selection procedure created a non-causal correlation between them.

% an example of causal sampling bias, where a sampling mechanism introduces a confounding correlation between multiple underlying causes in your data. Importantly, a causal factor (e.g., distance) is only considered to be confounding if (1) this factor does not actually cause the target outcome (e.g., cats versus dogs) and (2) is easier-to-compute than the desired causal factor. 

% The above scenario is an example of collider bias [CITE HERE], for which 

\begin{figure}[!htbp]
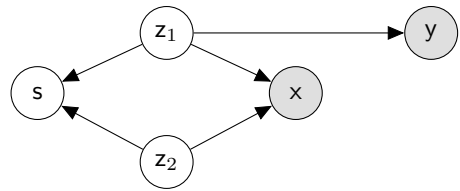

  \centering
  \tikz{ %
    \node[latent] (z1) {$\zRV_1$} ; %
    \node[latent, below=of z1] (z2) {$\zRV_2$} ; %
    \node[obs, right=of z1, yshift=-0.8cm] (x) {$\xRV$} ; %
    \node[obs, right=of z1, xshift=1.8cm] (y) {$\yRV$} ; %
    \node[latent, left=of z1, yshift=-0.8cm] (s) {$\sRV$} ; %
    \edge {z1, z2} {x} ; %
    \edge {z1} {y} ; %
    \edge {z1,z2} {s} ; %
  }
%   \vspace{3mm}
  
%     \tikz{ %
%     \node[latent] (z1) {$\zRV_1$} ; %
%     \node[latent, below=of z1] (z2) {$\zRV_2$} ; %
%     \node[obs, right=of z1, yshift=-0.8cm] (x) {$\xRV$} ; %
%     \node[obs, right=of z1, xshift=1.8cm] (y) {$\yRV$} ; %
%     \node[latent, left=of z1, yshift=-0.8cm] (s) {$\sRV$} ; %
%     \edge {z1, z2} {x} ; %
%     \edge {z1} {y} ; %
%     \edge {s} {z1} ; %
%   \edge {s} {z2} ; %
%   }
  \caption{Graphical model for collider bias. $\zRV_1$, $\zRV_2$ are latent information sources. The target $y$ is causally dependent on $\zRV_1$ but not $\zRV_2$. But $\xRV$ contains information from both latent sources. $s$ is a binary sample selection variable: $s=1$ implies that the sample is selected for a particular dataset.
%   \textbf{Top: collider bias} - training data is sampled from $P(\xRV,y|s=1)$ (the conditioning produces dependents between the parents of $s$), and test scenario from $P(\xRV,y)$. \textbf{Bottom: confounder bias} - training data is sampled from $P(\xRV,y)$, and test scenario from $P(\xRV,y|s=1)$ (the conditioning removes dependence between the child nodes of $s$).
  }\label{fig:GM}
\end{figure}

Consider an underlying data generating distribution $p_{\mathcal{D}}(\xRV, \yRV)$ for the data, $\mathcal{D}$. While for most problems we do not have access to this distribution, representative empirical samples are typically available. The bias problems we are concerned with are ones where there are multiple underlying variables, e.g.\ $\zRV_1$ and $\zRV_2$ in Figure \ref{fig:GM}, that causally influence the observed data.

When selecting or collecting data for a problem we inevitable introducing a \emph{sampling bias}, $\sRV$, which can be thought of as a form of rejection sampling. E.g., all the times a user implicitly chooses \emph{not} to take a photo. In this way sampling bias couples the underlying variables of the observed data in a manner that DNNs are ill-equipped to handle.
% such that: (1) $\xRV$ becomes a collider when the coupling occurs in the training data; (2) $\zRV_1$ and $\zRV_2$ have a non-causal association; and (3) the target variable, $\yRV$ remains caused by a single unobserved variable (animal type, for example). 

% Figure \ref{fig:GM}is a graphical model describing the relationship between observed data ($\xRV$ and $\yRV$ -- photos and labels, as per our example), the underlying causes (two, $\zRV_1$ and $\zRV_2$ -- animal type and distance), and the sampling mechanism, $\sRV$. While $p(\zRV_1, \zRV_2)$ is the distribution over all possible cause combinations, conditioning on the sampling mechanism -- $p(\zRV_1, \zRV_2 \mid \sRV)$ -- couples the underlying causes. Further, $\zRV_1$ and $\zRV_2$ are separate causes influencers of $\xRV$ but only $\zRV1$ influences $\yRV$, as per the desired objective.

We have included the sampling mechanism, $\sRV$, explicitly in Figure \ref{fig:GM}. Conditioning on $\sRV$ is what actually causes the association between underlying variables. In both cases the secondary signal does not cause the target and is therefore a \textbf{confounding signal}.
% \begin{enumerate}
%     \item $\sRV$ takes on different values during training and testing/use: $p(\zRV_1, \zRV_2 \mid \sRV={train}) \neq p(\zRV_1, \zRV_2 \mid \sRV={test})$. In this setting where $\zRV_1$ and $\zRV_2$ are non-causally associated via sampling, $\zRV_2$ can become a confounder of $\yRV$.
%     \item $\zRV_2$ {causes sufficiently informative cues} in $\xRV$ to predict $\yRV$.
% \end{enumerate}

\subsection{Confounding signals have high-gradient learning signals}\label{sec:versus}

Crucial to our insight into the tendency of DNNs to rely on easy-to-learn confounding signals in the training data is understanding how gradient of these signals with respect to the training loss evolves over learning. To this end we present a toy problem, called the one-pixel problem \footnote{Private communication: Harri Edwards, 2016.} in Section \ref{sec:opp}, and track the gradient ratio ($GR$) to determine the relative reliance on causal ($\zRV_{+}$) and confounding ($\zRV_{-}$) variables:

% While the notion of a bias being ``easier-to-compute'' than the true signal is useful, we need a demonstrable definition of easiness. We define the relative ease of computation between two variables, $\zRV_1$ and $\zRV_2$, in terms of the ratio of the L2 norm of their gradients with respect to training loss:

\begin{equation}
    GR({\zRV_{+}},{\zRV_{-}}) = \left\Vert \frac{\partial L}{\partial {\zRV_{+}}} \right\Vert_{2}\bigg/ \left\Vert\frac{\partial L}{\partial \zRV_{-}}\right\Vert_2,
    \label{eq:gr}
\end{equation}
where  $\left\Vert \cdot\right\Vert_2$ is the L2 Norm. When $GR = 1$ the learning process is not favouring either signal. When $GR < 1$ the learning process is favouring the confounding signal and when $GR > 1$ it is favouring the causal signal. As this quantity is impossible to compute when we do not have direct access to the causal variable, a toy problem enables us to assess this.

\subsection{One-Pixel problem}\label{sec:opp}
\begin{figure*}[!htbp]
\begin{center}

\begin{subfigure}{0.8\textwidth}
\includegraphics[width=0.98\linewidth]{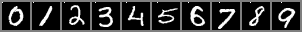}
\caption{Training images}
\end{subfigure}\\
\begin{subfigure}{0.25\textwidth}
\includegraphics[width=0.98\linewidth]{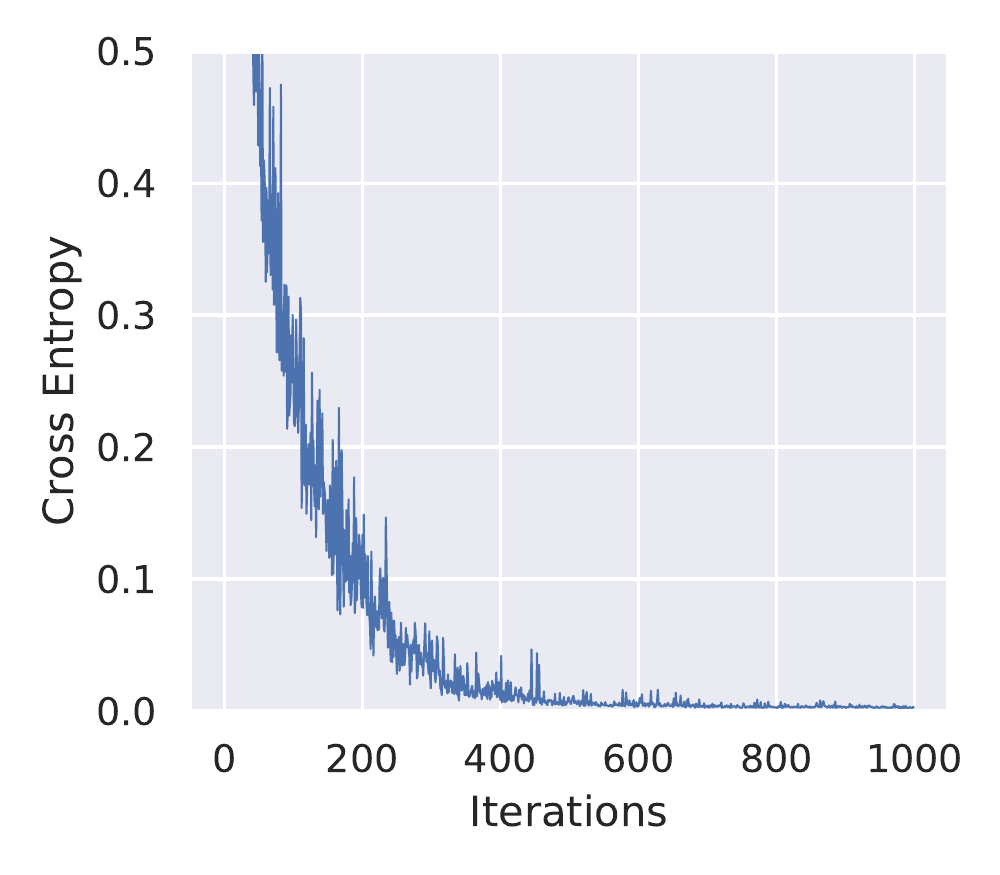}
\caption{Cross entropy loss}
\end{subfigure}%`   
\begin{subfigure}{0.25\textwidth}
\includegraphics[width=0.98\linewidth]{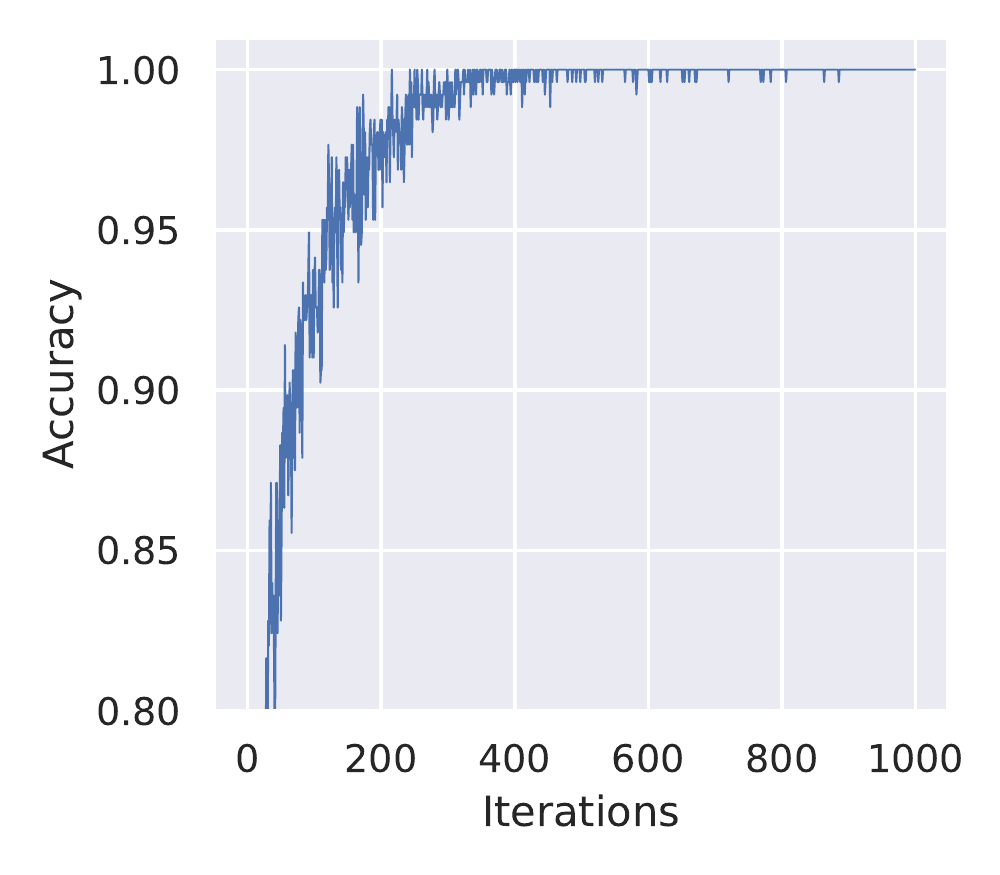}
\caption{Training accuracy}
\end{subfigure}%
\begin{subfigure}{0.25\textwidth}
\includegraphics[width=0.98\linewidth]{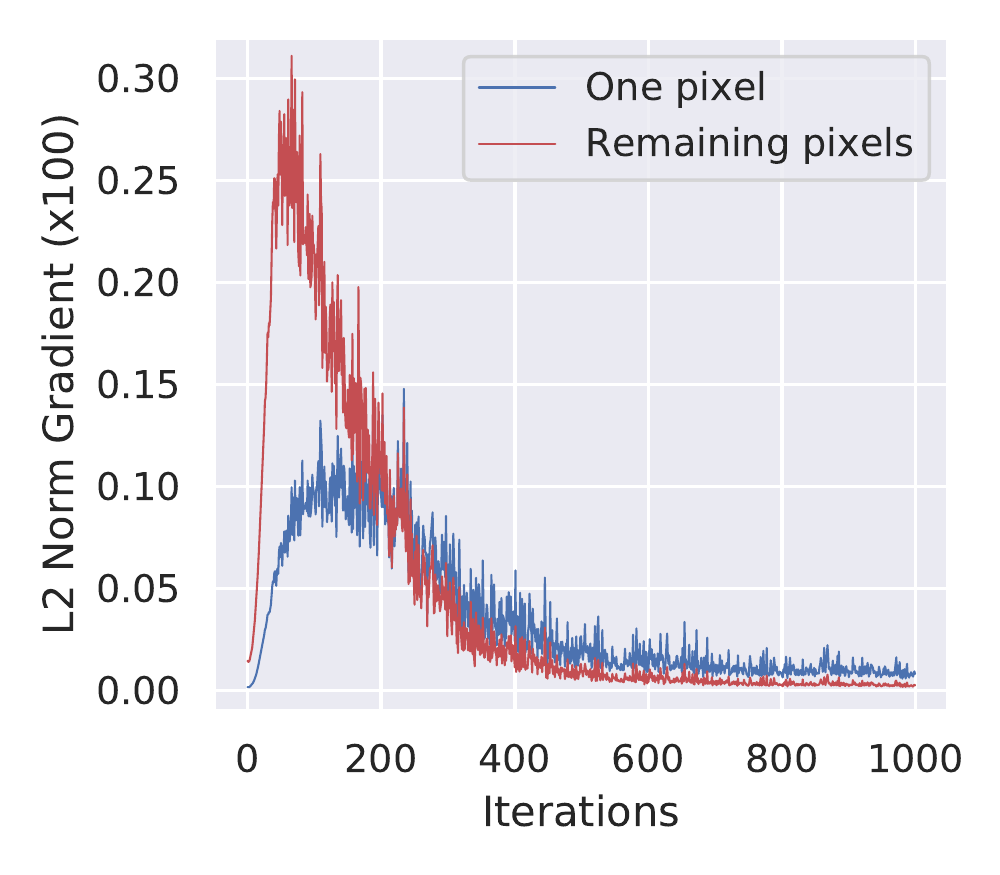}
\caption{L2 norm gradients}
\end{subfigure}%
\begin{subfigure}{0.25\textwidth}
\includegraphics[width=0.98\linewidth]{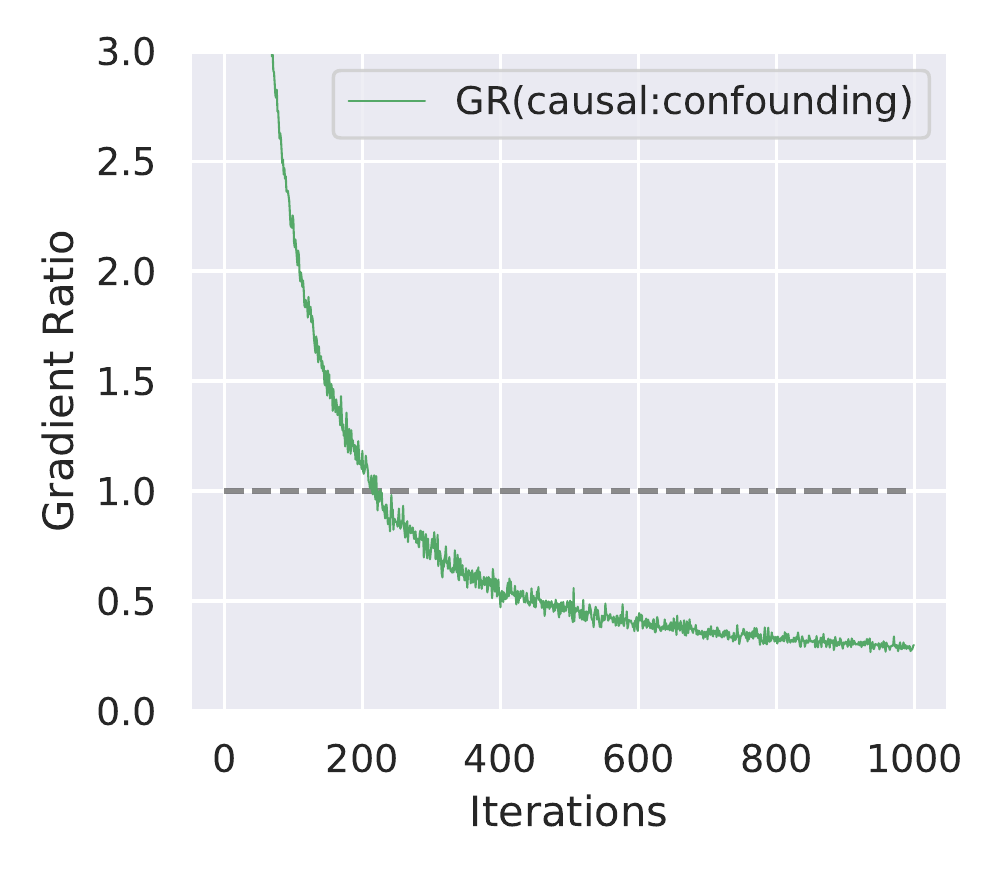}
\caption{Signal:bias gradient ratio, Equation \ref{eq:gr}}
\end{subfigure}

\end{center}
  \caption{One-pixel problem on MNIST. The class information is encoded as a confounding signal in the first $k$ pixels of the first row in each image (a). We show the training loss (b), training accuracy (b), l2 norm of gradients on the informative pixel and the remaining image (c), and the ratio of said gradients (d). In this case the test accuracy is no better than random chance. }
\label{fig:oppdemo}
\end{figure*}

The one-pixel problem is a bias-reliance demonstration that can be constructed simply using any image dataset for classification: all that it requires is setting the $k^{th}$ pixel of the first row of each training image to a pre-selected value (where $k$ is the class index) -- see Figure \ref{fig:oppdemo}. 

Figure \ref{fig:oppdemo} tracks the gradient ratio between causal signal and confounding signal, $GR({\zRV_{+}},{\zRV_{-}})$, measured as the ratio of L2 norm gradients on the image pixels (all but the $k$ pixels of the OPP) and the L2 norm gradients of the $k$ pixels encoding class information, over 1000 minibatch iterations. This is simple to compute for the one-pixel problem as there are distinct pixels associated with each signal. At initialisation the gradients are stronger over the image space (causal), but become dominated by the gradients over the $k$-pixel space (confounding) after only 230 iterations. This demonstration serves to show that neural networks prefer easy-to-compute confounder signals and that gradient information is the mechanism that preference takes.

\section{Prior Work}\label{sec:prior}

% mccoy2019
 DNNs lack robustness to human imperceptible perturbation called adversarial examples~\citep{goodfellow_explaining_2015, madry_towards_2018},  other semantic transformations such as rotations or translations~\citep{kanbak_geometric_2018,engstrom_rotation_2017,hendrycks_benchmarking_2019}, and more broadly to a domain shift in the input distribution~\citep{gulrajani_search_2020}.

Most relevant to our work is the observation that DNNs tend to rely on easy-to-learn biases or features that do not generalise outside the training distribution~\citep{geirhos_shortcut_2020}. In image classification, this is exemplified by the reliance of DNNs on the high frequency components in the image~\citep{jo_measuring_2017}. In natural language processing, DNNs were shown to latch onto statistical cues such as the present or absence of individual words~\citep{mccoy_right_2019}.

% \paragraph{Methods to robustify deep neural networks} 

Adversarial learning is a set of techniques particularly useful for improving the robustness of deep neural networks~\citep{goodfellow_explaining_2015,madry_towards_2018}. Predominantly, prior work focused on applying adversarial learning to improving the robustness to small perturbations bounded in $\ell_p$ norm. 

Adversarial learning was also applied to de-biasing models ~\citep{goel_model_2020, qiu_semanticadv_2019,zhang_mitigating_2018,beutel_data_2017,edwards_censoring_2015,arjovsky_invariant_2019,stachura_leakage-robust_2020}. \citet{zhang_mitigating_2018,beutel_data_2017,edwards_censoring_2015} remove the information about the bias from the input or hidden representation. \citet{arjovsky_invariant_2019} use adversarial learning to reduce reliance of a neural network on features that change between different domains. In contrast, we do not require annotated data on the presence of a bias. 

Data augmentation has also been shown to be effective in improving robustness of deep neural networks to semantically meaningful perturbations~\citep{fawzi_adaptive_2016,hendrycks_augmix_2020}. Our work can be seen as an automatic method for creating such augmentations.

Our work is most closely related to unsupervised methods to de-biasing models~\citep{nam_learning_2020,bras_adversarial_2020,gowal_achieving_2020,bahng_learning_2020}, but substantially differ in the assumptions made. \citet{gowal_achieving_2020} assume access to a disentangled representation with an identified factor that is not causally related to the label. Then they use a decoder to produce augmented images by mixing the spurious factor between different pairs of images. Specifically, they train StyleGAN~\citep{karras_style-based_2018}, and use its first stage representation as the spurious factor. Their method is hence limited to image classification scenarios in which features identified by StyleGAN correspond to the learned biases. \citet{bahng_learning_2020} require providing a biased model that heavily relies on the bias information in the dataset (e.g. a CNN that has small receptive fields that bias the model towards textural information). They train a debiased classifier by regularizing its representation to be statistically independent from the representation learned by the biased model. 

Similarly to \citet{nam_learning_2020} and \citet{bras_adversarial_2020}, LAD hinges on a relaxed assumption that the spurious correlation is easier to learn than the true signal. \citet{nam_learning_2020} assumes first examples that are learned are biased, and trains a second network that has an intentionally high loss on them. \citet{bras_adversarial_2020} filters out examples that can be classified using a simple linear model. These approaches are inspired by the phenomenon that DNNs first prioritise learning a consistent subset of \emph{easy-to-learn} examples~\citep{swayamdipta_dataset_2020,arpit_closer_2017}. 

In contrast to \citet{nam_learning_2020,bras_adversarial_2020} we do not assume that a subset of examples is free of bias. Instead, we modify all examples using carefully crafted adversarial examples to reduce the reliance on confounding signals. We compare directly to \citet{nam_learning_2020} and \citet{bahng_learning_2020} and show markedly improved generalisation performance. 

Finally, LAD is related to \cite{minderer_automatic_2020}. Similarly to us, they reduce the reliance of training on easy to learn features by performing an adversarial walk in the latent space of an autoencoder. The key difference is that they use the method to improve self-supervised learning, in which they note the self-superivsed objective can be too easily optimized by relying on shortcut (easy to learn) features.

% For example, \cite{arpit2017} show that in a dataset with some examples mislabeled , training priotized learning on clean examples. 

% Our work is most closely related to label-free methods to de-biasing models~\citep{nam2020learning}. Similarly to \citet{nam2020learning}, our method hinges on the assumption that the spurious correlation is easier to learn than the true signal. 

% https://arxiv.org/pdf/2009.10795.pdf nice references.

% \subsection{Paper structure}

% In this paper we present an algorithm that leverages the notion that biases are easier-to-compute than signals. Based on adversarial robustness training, our method (detailed in Section \ref{sec:method}) is designed to obscure bias-related manifestations in the input data ($\xRV$) such that their confounding correlations with the target ($\yRV$) is minimised. The effect is specific alteration to the data such that the model learned must rely more on the true signal. We present a number of toy and real-world experiments in Section \ref{sec:experiments} to show the efficacy of our method and discuss the findings in Section \ref{sec:discuss}.

% TODO: line of thought: stan high frequency components learned first. Cifar 10 corruptions break this.

\section{LAD: Latent Adversarial Debiasing}\label{sec:method}

We propose to counter collider bias in DNNs by augmenting the training data to artificially disassociate confounding signals and causal signals. We require three components to achieve this:

\begin{itemize}
    \item A latent representation, $\hRV$, of the underlying data manifold, modelled by $g$, such that the confounding and causal signals are approximately disentangled and accessible.
    \item A biased classifier, $f$, trained on this latent representation.
    \item A method to remove confounding signals from training examples using both $f$ and $g$.
\end{itemize}

 In the rest of this section we describe each component. We note here that similar but more restrictive assumptions were made by prior works~\citep{nam_learning_2020,bras_adversarial_2020,bahng_learning_2020,karras_style-based_2018}. Perhaps most importantly, we assume $f$ relies on the easy-to-learn confounding signal when trained on $\hRV$. We also constrain $f$ to ensure this is the case.
 
\subsection{Manifold Access}

We assume the data distribution is conditional on a set of underlying variables such that: $p(\xRV \mid \zRV_{1}, \zRV_{2})$ and $p(y \mid \zRV_{1})$ (see Figure \ref{fig:GM}.) This corresponds to an underlying low-dimensional manifold that dictates the space of plausible images in the data. 

We need access to (an approximation of) the latent manifold parameterised in a way that the easy-to-compute confounding signal is disentangled from the causal signal. This helps ensure that when we train a classifier $f$, it learns to rely on the information that induces bias, and allows us to alter or remove this information.

\citet{stutz_disentangling_2019} demonstrated a means of producing on-manifold adversarial examples by training class-specific variational autoencoder generative adversarial network (VAEGAN) hybrid models \citep{larsen_autoencoding_2016} for each class in the data. Via an adversarial walk on the approximated manifold space, \citet{stutz_disentangling_2019} were able to generate adversarial images with plausible deviations from the originals.

\paragraph{VQ-VAE: quantised latent space}
To satisfy the above desiderata, we use a vector quantised-variational autoencoder (VQ-VAE)  \citep{van_den_oord_neural_2017}. This model enables learning of a discrete (quantised) latent representation, where the number and size of the discrete codes are pre-chosen. Where \citet{stutz_disentangling_2019} learned VAEGAN models for each class to constrain the changes to remain on-manifold, the quantisation constraint of VQ-VAE offers a similar effect. We use the quantisation mechanism directly in the latent adversarial walk to project gradient-based changes onto the manifold. Early experimentation with standard autoencoders and VAEs evidenced that a strong constraint in the adversarial walk was paramount. 
% By quantising the latent codes, we reduce the likelihood of leaking information about the bias information. 

\begin{figure*}[!htbp]
\begin{center}
{\includegraphics[width=0.9\linewidth]{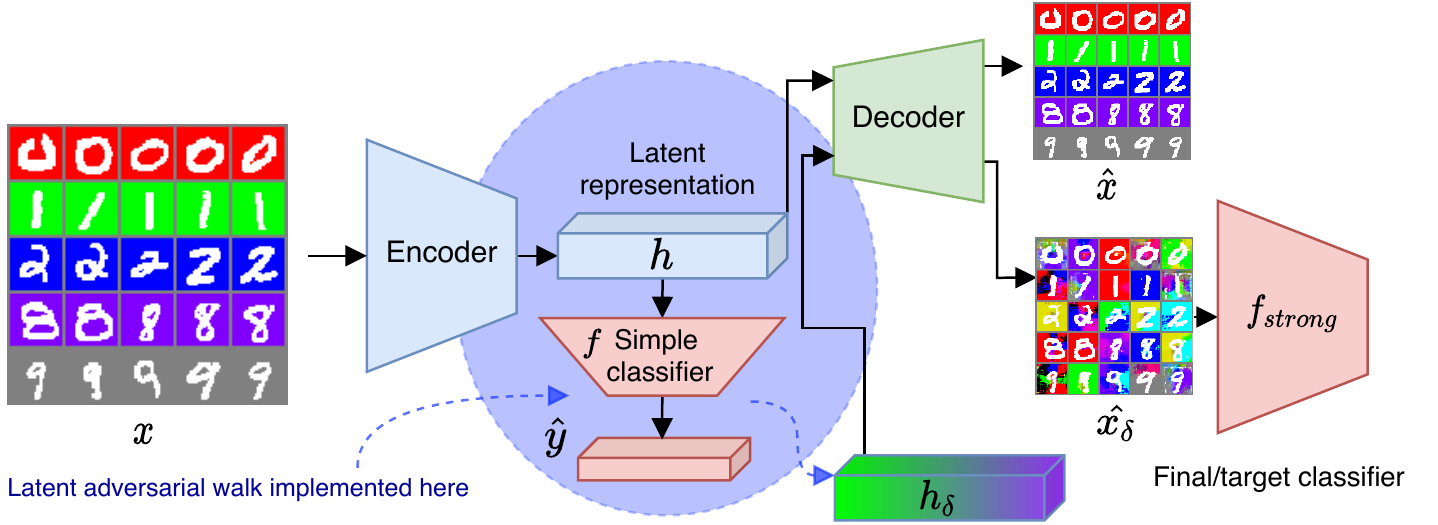}}

\end{center}
  \caption{Model setup. The VQ-VAE is the encoder decoder structure, tasked with learning quantised latent representation, $h$. The `simple' classifier, $f$, is learned using $(h, y)$ as input-target pairs, where $y$ is the class label. The latent adversarial walk (blue circle) alters $h$ such that $f$ produces high entropy (maximally uncertain) class predictions. The altered representation, $h_{\delta}$, can then be decoded into $x_{\delta}$ where the easy-to-learn confounding signal and causal signal are decoupled. The ultimate aim is to train an additional classifier, $f_{strong}$ using this augmented data for improved generalisation in collider bias settings.}
\label{fig:diagram}
\end{figure*}

The VQ-VAE is effectively an encoder decoder structure (See Figure~\ref{fig:diagram}) with a quantised latent space. Consider that the original image can be reconstructed as:
\begin{equation}
\begin{split}
    \hat{x} &= g(x)\\
     &= \text{dec}(h), h = \text{enc}(x),
\end{split}
\end{equation}
where $x$ and $\hat x$ are the input and reconstructed images, respectively, $g(\cdot)$ is the VQ-VAE, (dec, enc) are the decoder and encoder components thereof, and $h$ is the latent representation for $x$

\paragraph{Classification from bias} We can then attach a classifier, $f$, to this latent space in order to approximate the decision boundary associated with the easy-to-learn confounding signal:
\begin{equation}
    \hat y = f(h),
\end{equation}
where $\hat y$ is a class prediction, and train it using standard SGD to minimise the cross entropy loss. While $h$ will contain both confounding and causal signals, it is the tendency of $f$ to latch onto easy-to-learn features that enables LAD to work. In the following section we discuss how we use these two models $f(\cdot)$ and $g(\cdot)$ to traverse the latent space, $h$, and augment the training data  such that an additional classifier must rely on the causal signal.

LAD bears resemblance to \cite{gowal_achieving_2020} who assume that the lowest level representation learned by StyleGAN is not causally linked to the label. We relax this assumption in the sense that we only require our latent variable model to disentangle the underlying causal variables. We will rely on a (simple) classifier ($f$) and a gradient-based latent adversarial walk to decouple the confounding and causal signals.

\subsection{Latent adversarial walk}\label{sec:walk}

While a number of adversarial attack example generation algorithms exist \cite{szegedy_intriguing_2013, madry_towards_2018, goodfellow_explaining_2014}, these aim at producing imperceptible changes to an image such that a target classifier makes an incorrect, yet often highly confident, prediction on that image. Consider the family of white-box adversarial attacks \cite{madry_towards_2018} that maximise the training loss:

\begin{equation}
    \underset{\delta}{max} \mathcal{L}_{ce} (f(x+\delta), y)
\end{equation}
where $\mathcal{L}_{ce}$ is the cross entropy loss and $\delta$ is computed via projected gradient descent. $\delta$ is constrained to ensure perceptual similarity between $x$ and $x + \delta$. This is an optimisation process on the image space: an \textbf{adversarial walk} to systematically alter the image and maximise the training loss. In our case, however, we perform the adversarial walk on the latent space with the objective of maximising the entropy of the predictive probability:
\begin{equation}
    \underset{\delta}{max} {H} (f(\text{quantise}(h+\delta, G))),
\end{equation}
where ${H}$ is the entropy of the class probabilities $f$ computes, the $\text{quantise}(\cdot)$ function is the quantisation mechanism of VQ-VAE with learned latent codes. We compute $\delta$ as the standardised partial gradient to preserve the strength of the changes for any example:
\begin{equation}
    \delta = \alpha \cdot \left(\frac{\frac{\partial {H}}{\partial {h}} - \text{mean}(\frac{\partial {H}}{\partial {h}})}{\text{std}(\frac{\partial {H}}{\partial {h}})}\right),
\end{equation}
where $\text{mean}(\cdot)$ and $\text{std}(\cdot)$ are the mean and standard deviation computed over the entire gradient vector for each image, and $\alpha$ is a hyper-parameter dictating the step size of the walk. Standardisation helps ensure the steps taken by the gradient walk are approximately equal in length.

Algorithm \ref{alg:walk} details the \textbf{quantisation-constrained entropy-targeted adversarial walk} used for LAD. The end-goal here is to produce an altered latent representation, $h_{\delta} = \text{advwalk}(h, , f)$.

\begin{algorithm}[!htbp]
\SetAlgoLined
 \SetKwInOut{Input}{Input}
    \SetKwInOut{Output}{Output}
    \Input{$f(\cdot)$, $y$, $h$, $\alpha$, $steps$, $G(\cdot)$}
    \Output{Adjusted latent representation, $h_{\delta}$}
 $h_{delta} \gets h$ 
 \tcp{initialise $h_{\delta}$ to $h$}
 \For{$i\gets1$ \KwTo $steps$}{
    $\hat{y}\gets f(h_{delta})$ \tcp{compute prediction}
    $H \gets entropy (\hat{y})$ \;
    backprop to maximise $H$\;
    compute $\delta$\;
    $h_{delta} \gets h_{delta} + \alpha \cdot \delta$ \;
    $h_{delta} \gets \text{quantise}(G, h_{delta})$ ;
    }
 \caption{Quantisation-constrained entropy-targeted adversarial walk.\label{alg:walk}}
\end{algorithm}

\subsection{Post-walk classification}\label{sec:post}
Once the latent representation has been adjusted such that the simple classifier, $f$, outputs high entropy probabilities (i.e., it is unsure), we can then decode the new representation using the VQ-VAE decoder to construct a new image:
\begin{equation}
    \hat{x_{\delta}} = \text{dec}(h_{\delta}).
\end{equation}
The goal is that this new image should contain very little information related to what $f$ used to classify. Since $f$ is constrained and we assume it will latch onto the easy-to-learn confounding signals, this gives us a way of intentionally augmenting data to remove confounders. We can then use $x_{\delta}$ to learn an additional classifier in a standard fashion. We call this final debiased classifier $f_{strong}$. 

\section{Experiments}\label{sec:experiments}
We explore three datasets of increasing difficulty to assess the relative merit of Latent Adversarial Debiasing. For comparison with earlier works, we test two variants of coloured MNIST \citep{lecun_mnist_2010} (background and foreground) in Sections \ref{sec:colouredmnist1} and \ref{sec:colouredmnist2}. Although seemingly similar, these two variants of MNIST differ in the level of entanglement between image shape and colour. For the background-coloured variant, the colour is largely independent of the image shape, while for the foreground-coloured variant, the colour always occurs with shape. Following \citet{nam_learning_2020}, we also consider the corrupted CIFAR-10 dataset \citep{krizhevsky_learning_2009}. 

Unlike earlier works, we chose to consider the circumstance where the different confounding signals are pervasive and therefore do not assume any training data is free from bias. We seek to work toward handling the fundamental issue of confounding, instead of designing a method that is able to leverage small amounts of confounder-free data. Since earlier works do not consider fully confounded training data we also compare on the settings they target, ensuring the proportion of biased data is listed consistently for earlier works. 

We consider two forms of assessment. The first we call the \textbf{Independent} settings, denoted `cross-bias' by \cite{bahng_learning_2020}, where the confounding signal is independent of the causal signal during test (e.g., colours are sampled independently at random during test for coloured MNIST datasets). The second we call \textbf{Conditioned} settings where the sample selection is changed so the confounding signal is held constant during the test (e.g., the original MNIST test set with black a background and white foreground). These cover two different test-scenarios where the confounding signal has no influence in the test setting.

\subsection{Implementational Details}\label{sec:expsetup}
For all datasets we use a ResNet-20 \citep{he_identity_2016} for the final $f_{strong}$ classifier and a single hidden-layer multi-layer perceptron (width 100 units) for $f$. For both MNIST datasets, the VQ-VAEs were trained with 20 latent codes (learned discrete quantisations) of length 64. For corrupted CIFAR-10 the VQ-VAE was trained with 2056 latent codes of length 64.

For background coloured MNIST we used 20 steps with $\alpha=0.1$ for the adversarial walk, for foreground coloured MNIST we used 7 steps with $\alpha=0.1$, and for corrupted CIFAR-10 we used 4 steps with $\alpha=0.07$. These were determined using a brief hyper-parameter search and cross validation. The results given in the following sections were computed on the held-out test data. We used random crops with padding size of 4 for all datasets, random affine transformations for MNIST datasets (with limits of: $rotation=15~degrees$, $scale=[0.8, 1.1]$, and $shear=15~degrees$), and random horizontal flips for corrupted CIFAR-10.

\begin{table*}
\centering
\begin{tabular}{ccccc}
\textbf{Dataset}            & \textbf{Method}  & \textbf{Bias ratio} & \textbf{Accuracy (independent)} & \textbf{Accuracy (conditioned)} \\
\hline
\multirow{4}{*}{BG coloured MNIST}        &   LAD    & $100\%  $    & \textbf{98.82 $\pm$ 0.039 \%}   & $95.35 \pm 0.32 \%$   \\
&   Vanilla & $100\%  $    &  $0.00 \pm 0.00\%$& $10.64 \pm 0.65\%$                      \\
\cline{2-5}
& ReBias  & $99\%  $     & $88.1\%  $            & -                    \\
& ReBias  & $99.9\%$     & $22.7\%    $          & -                    \\

\hline
\multirow{5}{*}{FG MNIST} & LAD    & $100\% $     & \textbf{98.86 $\pm$ 0.10\% }    & $98.34 \pm 0.40\%$     \\
& Vanilla & $100\% $     &$0.01 \pm 0.01\%$ &  9.90 +- 0.13\\
\cline{2-5}
& LfF     & $95\% $      & $85.39 \pm 0.94\%$     & -                    \\
& LfF     & $99\% $      & $74.01 \pm 2.21\%$     & -                    \\
& LfF     & $99.5\% $    & $63.39\pm1.97\%  $     & -                    \\

\hline
\multirow{5}{*}{Corrupted CIFAR-10}& LAD    & $100\%$      & \textbf{39.93 $\pm$ 0.62\%}     & $51.89 \pm 0.32\%$     \\
& Vanilla & $100\%$      &$12.16 \pm 0.16$&  $21.49 \pm 0.16$\\
\cline{2-5}
& LfF     & $95\%$       &\textbf{ 59.95}$\pm$\textbf{0.16}\%       & -                    \\
& LfF     & $99\%$       & $41.37\pm2.34\%$       & -                    \\
& LfF     & $99.5\%$     & $31.66\pm1.18\%$       & -                    \\

\hline
\end{tabular}
\caption{Test accuracy on all datasets for two test conditions: the independent case and the conditioned case. The vanilla method is simply a $f_{strong}$ (c.f. Figure~\ref{fig:GM}) model trained on the original data. While we include results from ReBias \citep{bahng_learning_2020} and LfF \citep{nam_learning_2020}, their methods are not directly comparable because they assume the training set contains some percentage of unbiased data.  \label{tab:results}}
\end{table*}

\subsection{Background coloured MNIST}\label{sec:colouredmnist1}

To produce this dataset we used the code provided \footnote{\url{https://github.com/clovaai/rebias}} by the authors of ReBias \citep{bahng_learning_2020} and compare to their results in Table \ref{tab:results}. It is evident that ReBias is strongly dependant on using a small portion of unbiased data: at a bias ratio of 99\%, they achieved 88.1\% test accuracy on an unbiased test set but only 22.6\% at a bias ratio of 99.9\%. LAD achieved 98.82\% on this dataset at 100\% bias ratio, approaching what is achievable on standard MNIST. 

The efficacy of LAD is clearly evident by these results. We also show the training data LAD reconstructions in Figure \ref{fig:images} (a) and (b). LAD is clearly able to augment the colour information such that it is decoupled from the classification targets.

% \begin{figure}
% \captionsetup[subfigure]{justification=centering}
% \begin{center}

% \begin{subfigure}{0.23\textwidth}
% \includegraphics[width=0.99\linewidth]{figures/colourmnist-x.png}
% \caption{$x$, original training data with class-coupled colour}
% \end{subfigure}%
% \begin{subfigure}{0.23\textwidth}
% \includegraphics[width=0.99\linewidth]{figures/colourmnist-xaltered.png}
% \caption{$\hat{x_{\delta}}$, reconstruction after latent adversarial walk}
% \end{subfigure}\\

% \end{center}
%   \caption{Foreground coloured MNIST examples for training data (a) and reconstruction using LAD.}
% \label{fig:colourmnist}
% \end{figure}

% \begin{figure}
% \captionsetup[subfigure]{justification=centering}
% \begin{center}

% \begin{subfigure}{0.23\textwidth}
% \includegraphics[width=0.99\linewidth]{figures/cifar-x.png}
% \caption{$x$, training data with class-coupled corruptions}
% \end{subfigure}%
% \begin{subfigure}{0.23\textwidth}
% \includegraphics[width=0.99\linewidth]{figures/cifar-xaltered.png}
% \caption{$\hat{x_{\delta}}$, reconstruction after latent adversarial walk}
% \end{subfigure}\\
% \end{center}
%   \caption{Corrupted CIFAR-10 examples for training data (a), and reconstructions using LAD (b).}
% \label{fig:cifar}
% \end{figure}

\subsection{Foreground coloured MNIST}\label{sec:colouredmnist2}
To generate this data we followed the protocol in LfF \citep{nam_learning_2020}. Ten colours were randomly chosen for each class. For each image an RGB colour is sampled (correlated with class for training) from $\mathcal{N}\sim(RGB, 0.005)$: Gaussian noise is added to the selected mean colour with a standard deviation of 0.005.

Similar to background coloured MNIST, LAD exceeds the current state-of-the-art, even though we consider 100\% biased training data: the closest comparison of LfF achieves a test accuracy of 63.39\% at 99.5\% bias while we achieve 98.86\% test accuracy. Again, these results are approaching what is achievable on the standard MNIST dataset. However, a close inspection of Figure \ref{fig:images} (c) and (d) will evidence that the alterations owing to LAD do begin to affect the the actual digit shape -- note particularly the nines in Figure \ref{fig:images} (d). Compare this to (b) where the altered data leaves the digit shape almost entirely unchanged. This difference is owing to the level of entanglement between bias and true signal. Foreground coloured MNIST has the bias variable (colour) overlayed on the true signal (digit shape) instead of as a static background. Nonetheless, the test accuracies are almost identical.

\subsection{Corrupted CIFAR-10}\label{sec:cifar}

Next we consider a constructed dataset where the bias and signal variables are far more entangled. Following \cite{nam_learning_2020}, we construct a variant of corrupted CIFAR-10 Corrupted CIFAR-10\textsuperscript{1} in \citet{nam_learning_2020}, with corruptions: Snow,
Frost, Fog, Brightness, Contrast, Spatter, Elastic, JPEG, Pixelate, Saturate. During training these correlate with each class. Using corruptions to benchmark neural network robustness is not new -- \cite{hendrycks_benchmarking_2019}, and using them as a class-informative bias yields an extremely challenging dataset. 

\begin{figure}
\captionsetup[subfigure]{justification=centering}
\begin{center}
\begin{subfigure}{0.23\textwidth}
\includegraphics[width=0.99\linewidth]{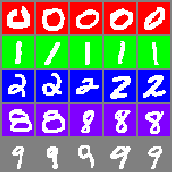}
\caption{$x$: BG coloured MNIST}
\end{subfigure}%
\begin{subfigure}{0.23\textwidth}
\includegraphics[width=0.99\linewidth]{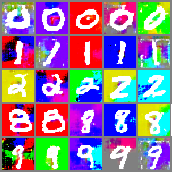}
\caption{$\hat{x_{\delta}}$: LAD reconstruction}
\end{subfigure}\\
\begin{subfigure}{0.23\textwidth}
\includegraphics[width=0.99\linewidth]{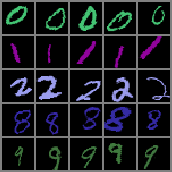}
\caption{$x$: FG coloured MNIST}
\end{subfigure}%
\begin{subfigure}{0.23\textwidth}
\includegraphics[width=0.99\linewidth]{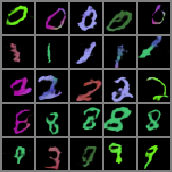}
\caption{$\hat{x_{\delta}}$: LAD reconstruction}
\end{subfigure}\\
\begin{subfigure}{0.23\textwidth}
\includegraphics[width=0.99\linewidth]{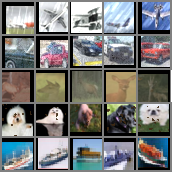}
\caption{$x$: corrupted CIFAR-10}
\end{subfigure}%
\begin{subfigure}{0.23\textwidth}
\includegraphics[width=0.99\linewidth]{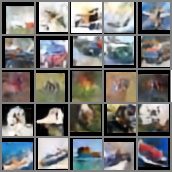}
\caption{$\hat{x_{\delta}}$: LAD reconstruction}
\end{subfigure}
\end{center}
  \caption{Training data examples for (a) background coloured MNIST, (c) foreground colorued MNIST, and (e) corrupted CIFAR-10, with corresponding LAD reconstructions for each dataset in (b), (d), and (f), respectively.}
\label{fig:images}
\end{figure}

Not only are corruptions often nuanced, they are also destructive, meaning that distentangling these from the underlying image is not always possible. Blurring, elastic distortion, JPEG compression, and pixelation are all examples non-reversible corruptions. While some (like contrast or brightness adjustment) are easier to change, it is understandable that earlier work was only able to achieve 31.66\% test accuracy at a high bias ratio. We were able to achieve 39.93\% test accuracy at a bias ratio of 100\%. While LfF can achieve 59.95\% on corrupted CIFAR-10, this requires a relatively low bias ratio of 95\%, once more evidencing the reliance of earlier works on bias-free training data. We also note that the reconstructions in Figure \ref{fig:images} (f) are blurry, highlighting the need for an improved encoder-decoder model.

% \subsection{Celeb Faces}
% A subset (I think) of CelebA faces to exemplify the bias hair colour/heavy makeup for class gender. 

% We also achieved substantial gains over the vanilla model in the confounder bias test setup (bias-free accuracy in Table \ref{tab:results}). This shows clearly that LAD can be used to mitigate both collider and confounder bias.
\section{Discussion and Conclusion}\label{sec:discuss}
DNNs tend to focus on easy-to-learn features, should those features be sufficiently informative of the target. In this paper we showed that this problematic behaviour means that DNNs are ill-equipped to handle a ubiquitous form of sample selection bias known as collider bias. The process of collecting and curating training data can often create a scenario where test data differs substantially from training data. When the training data contains a confounding signal (such as lighting conditions), DNNs will generalise poorly. We argue that it is the deep structure of neural networks, combined with the gradient-driven learning process used that amplifies their dependence on easy-to-learn confounding signals.

We presented LAD, a method to produce latent adversarial examples that specifically target the easy-to-learn confounding signals in the data manifold. Using a VQ-VAE to approximate the data manifold corresponding to causal and confounding signals, we leverage the tendency of DNNs to latch on to easy-to-learn features and define an appropriate adversarial walk on this manifold. Decoding the adjusted latent manifold back to the image space yields augmented data where confounding signals are largely mitigated against. A classifier trained on this new data generalises better. We evidenced substantial test accuracy gains of 76.12\% on background coloured MNIST, 35.47\% on foreground coloured MNIST, and 8.27\% on corrupted CIFAR-10, even when the training data was 100\% biased. 

Since LAD does not require any bias-free data, we believe we are moving toward solving the broad issue that neural networks latch on to easier-to-learn features, and evidence this in the collider bias settings. While we focus on constructed datasets that demonstrate effectively the problem at hand, extending the ideas and solutions presented herein to broader notions of dataset bias and neural network robustness is a natural progression and is planned for future work. 
% While the experiments presented in this paper are specific to a situation where the training data is a small subset sampled from a broader data generating distribution -- a subset where an additional bias is informative of class -- extending the ideas presented in here to broader notions of dataset bias and neural network robustness is a natural progression and is planned for future work. 
% \subsubsection*{References}
\subsubsection*{Acknowledgements}
Our work was supported in part by the EPSRC Centre for Doctoral Training in Data Science, funded by the UK Engineering and Physical Sciences Research Council (grant EP/L016427/1) and the University of Edinburgh. The opinions expressed and arguments employed herein do not necessarily reflect the official views of these funding bodies.
\bibliography{references}
\bibliographystyle{plainnat}

% References follow the acknowledgements.  Use an unnumbered third level
% heading for the references section.  Please use the same font
% size for references as for the body of the paper---remember that
% references do not count against your page length total.

% \begin{thebibliography}{}
% \setlength{\itemindent}{-\leftmargin}
% \makeatletter\renewcommand{\@biblabel}[1]{}\makeatother
% \bibitem{} J.~Alspector, B.~Gupta, and R.~B.~Allen (1989).
%     \newblock Performance of a stochastic learning microchip.
%     \newblock In D. S. Touretzky (ed.),
%     \textit{Advances in Neural Information Processing Systems 1}, 748--760.
%     San Mateo, Calif.: Morgan Kaufmann.

% \bibitem{} F.~Rosenblatt (1962).
%     \newblock \textit{Principles of Neurodynamics.}
%     \newblock Washington, D.C.: Spartan Books.

% \bibitem{} G.~Tesauro (1989).
%     \newblock Neurogammon wins computer Olympiad.
%     \newblock \textit{Neural Computation} \textbf{1}(3):321--323.
% \end{thebibliography}

% \bibliography{references, covariateshift}

\end{document}